\documentclass{article} 
\usepackage[final]{colm2026_conference}

\usepackage[utf8]{inputenc}
\usepackage[T1]{fontenc}
\usepackage{microtype}
\usepackage{graphicx}
\usepackage{float}
\usepackage{booktabs}
\usepackage{amsfonts}
\usepackage{amsmath}
\usepackage{amssymb}
\usepackage{nicefrac}
\usepackage{xcolor}
\usepackage{multirow}
\usepackage{subcaption}
\usepackage{hyperref}
\usepackage{url}
\usepackage{lineno}
\usepackage{mathtools}
\usepackage{amsthm}
\usepackage{iftex}

\definecolor{darkblue}{rgb}{0, 0, 0.5}
\hypersetup{colorlinks=true, citecolor=darkblue, linkcolor=darkblue, urlcolor=darkblue}
\hypersetup{
  pdftitle={Plans Fade Fast: Why Context Management Is Load-Bearing for LLM Agents},
  pdfauthor={Aman Mehta and Anupam Datta},
  pdfkeywords={LLM agents, context management, plan persistence, replay pairing}
}

\title{Plans Fade Fast:\\Why Context Management Is Load-Bearing for LLM Agents}

\author{Aman Mehta \\
Snowflake AI Research \\
\texttt{aman.mehta@snowflake.com}
\And
Anupam Datta \\
Snowflake AI Research \\
\texttt{anupam.datta@snowflake.com}}

\begin{document}

\ifcolmsubmission
\linenumbers
\fi

\maketitle

\begin{abstract}
Context managers summarize or evict old messages so agents can operate beyond a finite window. This creates a reliability problem: removing history that still shapes the model's computation may change later actions, while retaining everything defeats compression. We study plans because agents create them early and are expected to use them later, yet behavioral success alone cannot reveal whether later calls still depend on the plan exchange. We make four contributions. First, we introduce \emph{replay pairing}, which holds the action-observation trajectory fixed and compares pre-action hidden states with and without the plan exchange. We show that, on Llama-3.1-70B in ALFWorld, the mean contrast across 15 sampled layers decreases $4.1\times$ from the first post-plan action to the second; Llama-3.1-8B shows a similar $5.6\times$ decrease. Second, we show that a simple linear model using layer-$L_{32}$ hidden states predicts later contrast on entirely held-out tasks better than a baseline that only knows the action-call offset ($R^2=0.847$ vs.\ $0.575$). Third, we identify confounds that limit this measurement: the paired histories differ in length, token positions, and discourse structure, and replayed reasoning can reintroduce plan-derived text. Reasoning traces therefore require symmetric controls, and native Qwen3 shows that the Llama timing pattern is not universal. Fourth, in a separate descriptive compression test, we find pooled ALFWorld run success of $56.7\%$ with full history and $22.0\%$ with naive eviction; preserving or re-surfacing only the plan does not recover the gap. Replay pairing measures sensitivity to retaining an exchange, not isolated plan content or later eviction. It gives researchers a concrete audit tool and suggests that context policies should preserve broader task state, not plans alone.
\end{abstract}

\section{Introduction}
\label{sec:intro}

We study a practical reliability problem in long-horizon agents. Agents often write a plan near the start of a task and then accumulate observations, actions, and tool outputs. A context manager must eventually summarize or evict older messages. Removing a still-influential plan can silently change later computation, while retaining all history defeats compression. Agent builders therefore need a way to test which parts of history still matter.

Existing behavioral metrics cannot answer this question. A correct action may follow from the latest observation rather than the old plan, while a wrong action does not reveal what was forgotten. Planning frameworks such as ReAct \citep{react}, Chain-of-Thought \citep{cot}, Tree of Thoughts \citep{yao2023tree}, and Reflexion \citep{shinn2023reflexion} make intermediate reasoning visible, but do not measure how strongly a later model call still depends on one earlier exchange.

We introduce \emph{replay pairing} as a context-sensitivity test (Figure~\ref{fig:replay}). Condition A retains the plan exchange. Condition B replays the same action-observation trajectory but omits that exchange. We compare their pre-action hidden states using cosine distance; this is the \emph{plan-exchange contrast}. A large contrast means that retaining the exchange changes the current computation. A small contrast means that the two replay states have become more similar.

We show that, on Llama-3.1-70B in ALFWorld \citep{alfworld}, the mean contrast across 15 sampled layers is $0.453$ at the first action after the plan exchange and $0.110$ at the second, a $4.1\times$ decrease. Llama-3.1-8B shows a similar $5.6\times$ decrease. Thus, in these Llama settings, the effect of retaining the exchange becomes much smaller after one shared action-observation cycle.

We emphasize that this contrast is broader than plan content. The two histories also differ in length, token positions, and discourse structure, and A's replayed thoughts can later reintroduce plan-derived text into B. Replay pairing is therefore not an eviction experiment and cannot show that a previously seen plan was forgotten. It measures the effect of retaining the whole exchange.

We next develop a simple diagnostic for the later contrast. We show that a linear model using the $L_{32}$ hidden state predicts contrast on completely held-out tasks better than a quadratic baseline that only knows the action-call offset (\S\ref{sec:detection}). This result shows that the hidden state contains predictive information beyond elapsed steps, although the layer was selected on the full dataset and the analysis remains exploratory.

\textbf{Boundary conditions.} We identify a reasoning-trace confound: prior reasoning can copy plan text into the stripped replay, so controls must transform both histories symmetrically. Native Qwen3-thinking shows an exploratory contrast that grows rather than decreases, demonstrating that the Llama timing pattern is not universal. Our primary evidence is representational and covers short text-based trajectories.

\textbf{Paper roadmap.} The paper follows three practical questions. First, how does sensitivity to the plan exchange change across action calls (\S\ref{sec:diagnose})? Second, can the current hidden state predict that later sensitivity beyond elapsed steps (\S\ref{sec:detection})? Third, what happens under actual context compression, and is preserving only the plan enough (\S\ref{sec:compression})? Sections~\ref{sec:crossdomain} and~\ref{sec:reasoning} state where the evidence does and does not generalize.

\textbf{Contributions.}
\begin{enumerate}
\itemsep0pt
\item \textbf{Measurement:} replay pairing tests how strongly a current model call depends on retaining one earlier exchange, while making its content, length, position, and discourse confounds explicit.
\item \textbf{Primary finding:} the contrast drops $4.1\times$ between the first and second post-plan actions in Llama-3.1-70B on ALFWorld; Llama-3.1-8B shows a similar $5.6\times$ decrease.
\item \textbf{Diagnostic:} a simple linear model using $L_{32}$ hidden states predicts later contrast on held-out tasks better than a quadratic action-offset baseline.
\item \textbf{Control requirement:} reasoning traces can contaminate the stripped replay, while transforming only one history creates a separate distribution shift.
\item \textbf{Operational lesson:} naive compression has a much lower descriptive success rate than full history, and preserving or re-surfacing only the plan does not recover the gap.
\end{enumerate}

\section{Related Work}
\label{sec:related}

\textbf{Context-time vs.\ weight-time memory.} A growing line of work asks where information used by a model resides: in \emph{context-time} memory (text supplied in the inference window) or \emph{weight-time} memory (information absorbed into parameters). Retrieval-head analysis shows that a small set of attention heads supports long-context factuality by copying from the window \citep{wu2024retrieval}. Multi-turn studies also find that models can lose track of information distributed across turns \citep{laban2025lostMulti}. Compression, summarization, and KV-cache eviction assume that some tokens can be removed without changing behavior. These results motivate context management, but do not provide a paired trajectory-level test of whether retaining one earlier exchange still changes current computation. Replay pairing fills that measurement gap; it does not determine whether a previously seen plan has been internalized.

\textbf{Planning and faithfulness in LLM agents.} The standard paradigm interleaves reasoning and acting \citep{react, cot, yao2023tree, shinn2023reflexion, huang2022inner}. These frameworks assume explicit plans improve behavior, which behavioral metrics can test, but they do not show how sensitivity to one plan exchange changes over later calls. Repeated-run studies show that multi-step agents can diverge early under identical inputs, and that repeated agreement can still preserve an incorrect interpretation \citep{mehta2026behavioral,mehta2026confident}. Cross-run hidden-state convergence at an early step predicts this behavioral consistency but not correctness \citep{mehta2026commit}. We add a representational comparison; we do not infer behavior from the contrast alone.

\textbf{Hidden-state representations.} LLM hidden states linearly encode many semantic properties \citep{burns2023discovering, marks2024geometry, zou2023repe}. Circuit tracing \citep{anthropic_circuits} and the Tuned Lens \citep{belrose2023eliciting, nostalgebraist2020logitlens} examine layers within one forward pass. Replay pairing instead follows a paired hidden-state contrast across agent calls.

\textbf{Interventions and reasoning-model probes.} Activation steering \citep{turner2023steering, zou2023repe}, inference-time intervention \citep{li2023inference}, and self-refinement \citep{madaan2023selfrefine} target hidden states or context. Several studies probe reasoning-model hidden states in single-turn settings \citep{zhang2025reasoning, zhang2025fromreasoning, vilas2026tracing}. \citet{kwon2026reasonif} study instruction following inside reasoning traces, \citet{wu2025thinkInter} propose Thinking Intervention, and \citet{stolfo2025steering} use activation differences for steering. We identify a contamination problem that affects paired-trajectory measurements in multi-turn reasoning agents and show why the controls must be symmetric.

\section{Measuring Sensitivity to a Plan Exchange}
\label{sec:method}

\subsection{Agent Setup}

We run the models as ReAct agents \citep{react}. The primary non-reasoning model is Llama-3.1-70B-Instruct \citep{llama3}, with Llama-3.1-8B-Instruct for a within-family replication. DeepSeek-R1-Distill-Llama-70B \citep{deepseek2025r1} is derived from Llama-3.3-70B-Instruct and fine-tuned on samples generated by DeepSeek-R1. It emits explicit \texttt{<think>}$\ldots$\texttt{</think>} blocks before each action. We also report an exploratory temporal pattern from native Qwen3-32B with thinking enabled \citep{qwen3}. For R1-Distill-Llama, we inject a \texttt{<think>$\backslash$n} prefix so the reasoning trace opens with the expected tag.

The primary environment is ALFWorld \citep{alfworld}, a text-based household environment where the agent completes tasks by navigating rooms and manipulating objects through a natural-language action space (\texttt{go\_to}, \texttt{pick\_up}, \texttt{put}, \texttt{heat}, \texttt{cool}, \texttt{clean}, \texttt{examine}). It spans six task types ranging from $3$-$4$ to $6$-$8$ plan steps. We use the text-only (non-visual) variant, following standard ReAct-on-ALFWorld baselines \citep{react}.

An archived exploratory collection covers 30 HotpotQA tasks \citep{hotpotqa} with two runs each. It uses an earlier runner whose second post-plan measurement is contaminated by replayed plan-bearing text. Section~\ref{sec:crossdomain} audits this defect; we do not use HotpotQA as validation evidence. Its plan request asks for the entities or documents to search, the required evidence, and how to combine it.

The primary 70B models run on $8\times$H200 GPUs with pipeline parallelism. Hidden states are captured at the last prompt token, after applying the chat template and adding the assistant-generation marker but before generating the action response. Their dimensions depend on the checkpoint: 8,192 for Llama-3.1-70B and R1-Distill-Llama-70B, 4,096 for Llama-3.1-8B, and 5,120 for Qwen3-32B.

\subsection{Plan Elicitation}
\label{sec:plan_elicit}

To study plan faithfulness we first need an explicit plan, so we inject a guard message asking the agent for a multi-step plan:

\begin{quote}
\small
\textit{``Before continuing, state your step-by-step plan: list each physical
action you will take in order (go to X, pick up Y, heat in Z, put on W).
Be specific about object names and locations from what you have observed.''}
\end{quote}

The agent replies with a sequential plan, which then stays in history for the rest of the task. For HotpotQA, the guard instead asks for the expected answer type, the entities to investigate, and the plan for finding the answer. We inject the guard before call 2 in ALFWorld and before call 4 in HotpotQA.

\subsection{Replay Pairing}
\label{sec:replay}

\begin{figure}[t]
\centering
\includegraphics[width=0.98\textwidth]{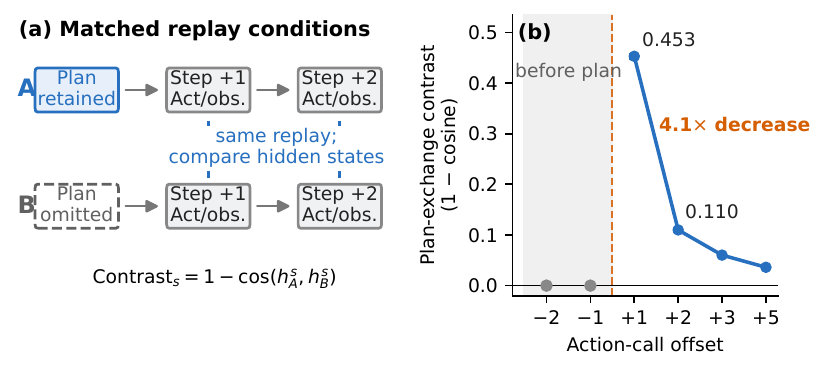}
  \caption{Replay pairing and measured decay. \textbf{Left:} conditions A and B replay the same action-observation trajectory but differ in whether the plan exchange is present. \textbf{Right:} the mean ALFWorld contrast drops $4.1\times$ from the first to the second action call after the plan exchange (error bars: mean per-layer SE).}
\label{fig:replay}
\end{figure}

For each primary ALFWorld run we create two matched conditions.

\textbf{Condition A (plan-present).} The agent runs normally with the guard before action call 2; the plan enters history. The extra call that generates the plan is not measured. We capture hidden states $\mathbf{h}_A^{(s,\ell)}$ before each normal action response.

\textbf{Condition B (plan-stripped replay).} We replay A's exact trajectory, including A's observations and full Thought:/Action: responses, but omit the guard exchange. We call the LLM on this history to produce $\mathbf{h}_B^{(s,\ell)}$ and discard B's output; A's trajectory drives the conversation.

Both conditions replay the same action-observation sequence; they differ in whether the plan exchange is present. The plan-exchange contrast at measured call $s$ and layer $\ell$ is:
\begin{equation}
  \text{PlanExchangeContrast}^{(s,\ell)} = 1 - \cos\!\left(\mathbf{h}_A^{(s,\ell)},\;
  \mathbf{h}_B^{(s,\ell)}\right)
\end{equation}
The plan-generation call is not captured. Action-call offset $+1$ is the first normal action call after the exchange, and $+2$ is the second; offsets $-2$ and $-1$ are pre-plan baselines. Conditions A and B are identical before the plan, and their contrast is below $10^{-8}$ (Appendix~\ref{app:probe_details}). The probe dataset excludes action-call offset $+1$. It calls the next state \emph{probe offset} $+1$, so probe offsets are one smaller than the action-call offsets in Figures~\ref{fig:replay} and~\ref{fig:decay_curve}.

\textbf{What the contrast does and does not measure.} A and B differ in plan content, history length, token positions, and discourse structure. After the first measured action, B also receives A's plan-conditioned Thought:/Action: response, which can reintroduce plan-derived text into the shared suffix. The pre-plan baseline checks the replay implementation but does not separate these factors. Length-matched filler, shuffled-plan placebos, paraphrases, and plan ablations are needed to isolate content. Confounds may amplify or cancel one another, so the contrast is not an upper bound on a content-specific effect.

\subsection{Strict Stripping for Reasoning Models}
\label{sec:strict}

Standard replay pairing attributes later between-condition differences to retaining versus omitting the original plan exchange, subject to the confounds above. Reasoning-model traces add another complication. As the model works through each subsequent action it re-states parts of the plan, both inside its \texttt{<think>}$\ldots$\texttt{</think>} block and, to a lesser degree, in the retained Thought:/Action: text. When those turns are replayed verbatim into condition B, that re-stated plan content reappears even though the original plan exchange was stripped, which can reduce the measured between-condition contrast. This does not identify a content-specific effect.

We first test \emph{asymmetric stripping}: in addition to removing the plan exchange from B's history, remove every complete \texttt{<think>}$\ldots$\texttt{</think>} block from each prior assistant turn in B. Content after \texttt{</think>} (Thought: and Action: lines) is kept so that B still sees the same external action-observation sequence as A. This archived run already has pre-plan $L_{32}$ contrast $0.0199$, directly showing that editing only B creates a non-plan distribution shift. A proper control must apply the same transformation to both histories.

\textbf{What the archive establishes.} Asymmetric stripping removes whole reasoning blocks rather than only plan-relevant spans and changes only B's local distribution. We therefore do not interpret its increase as recovery. Five archived symmetric A--B pairs have zero pre-plan difference and are consistent with applying the transform to both histories, but their collection runner and seed were not retained. We treat their post-plan contrast as descriptive, not as a validated recovery estimate. Whole-block deletion also does not isolate plan content. Length-matched replacement and plan-span redaction remain open controls.

\subsection{Probe and Task Partitioning}
\label{sec:probe_arch}

We train a Ridge regression probe on condition-A hidden states at layer $\ell^*=32$ (Section~\ref{sec:diagnose}). The target is the continuous layer-specific contrast $y^{(s)}=\text{PlanExchangeContrast}^{(s,\ell^*)}$. The step-only baseline uses probe offset and its square. The primary probe data contain 80 ALFWorld tasks. Because $L_{32}$ was selected from the full ALFWorld dataset rather than nested within each fold, the readout remains exploratory.

\section{Finding 1: The Contrast Shrinks After One Action}
\label{sec:diagnose}

We first quantify how the effect of retaining the plan exchange changes once both histories accumulate the same action-observation suffix.

\subsection{First-to-Second-Action Decrease}

\begin{figure}[H]
  \centering
  \includegraphics[width=0.68\linewidth]{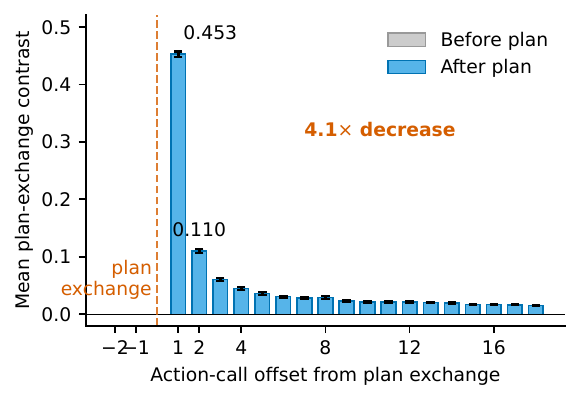}
  \caption{Mean plan-exchange contrast across 15 sampled layers (80 tasks; error bars are mean per-layer SE). The contrast drops $4.1\times$ from the first to the second action call after the plan exchange.}
  \label{fig:decay_curve}
\end{figure}

Figure~\ref{fig:decay_curve} shows the mean plan-exchange contrast across 15 sampled layers on ALFWorld (80 tasks). The pre-plan contrast is below $10^{-8}$, which validates the replay implementation. We find a mean contrast of $0.453$ at the first action call after the plan exchange and $0.110$ at the second, a $4.1\times$ decrease.

\subsection{Layer and Task-Type Profile}

We find that the first-action contrast peaks at $L_{32}$ ($0.686\pm0.024$), with a broad mid-network band around it. We use $L_{32}$ as $\ell^*$. Averaged across layers, all six ALFWorld task types have similar first-action magnitudes ($0.445$-$0.474$) and first-to-second-action decreases ($4.0$-$4.4\times$). The pattern is also similar for elicited plans ranging from 3--4 to 6--8 listed actions (Appendix~\ref{app:layer}).

\section{Finding 2: Hidden States Predict the Later Contrast}
\label{sec:detection}

Contrast is strongly related to action offset, so a predictor can appear useful simply by learning how much time has passed. We therefore test a stricter claim: hidden states predict variation in the contrast on entirely unseen tasks beyond what action offset already explains.

\subsection{Held-Out-Task Prediction}

\begin{figure}[H]
  \centering
  \includegraphics[width=0.98\linewidth]{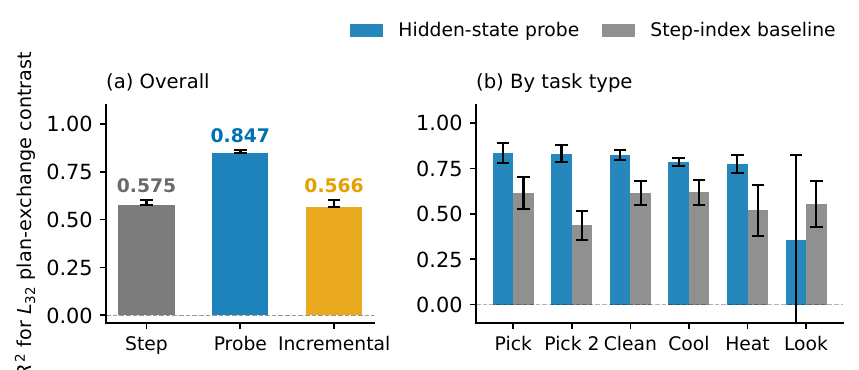}
  \caption{Task-grouped regression (mean $\pm$ SD across folds). \textbf{Left:} the hidden-state probe predicts the $L_{32}$ contrast better than a step-only baseline and explains additional residual variance. \textbf{Right:} the same comparison by ALFWorld task type; task counts from left to right are 19, 15, 18, 10, 13, and 5. The five-task Look-in-Light subset is the exception.}
  \label{fig:task_grouped}
\end{figure}

At $L_{32}$, we train a Ridge probe to predict the layer-specific contrast from 1,360 post-plan action states across 80 ALFWorld tasks. We evaluate it with five task-grouped folds, each holding out 16 complete tasks. The probe reaches mean $R^2=0.847\pm0.018$ across folds, compared with $0.575\pm0.025$ for the step-only baseline. After residualizing that baseline, the probe explains partial $R^2=0.566\pm0.039$ (Figure~\ref{fig:task_grouped}; Appendix~\ref{app:leakage}). Reported variation is SD across folds.

\subsection{What the Probe Does and Does Not Show}
\label{sec:probe_validity}

The task-grouped regression above is the primary probe result. Appendix~\ref{app:leakage} reports additional exploratory diagnostics based on probe offsets: shuffled labels, step-index decoding, and a four-class confusion matrix. These diagnostics are descriptive and do not establish a deployable detector.

\subsection{Observed-Contrast Lag Analysis}
\label{sec:lag}

This analysis does not use the trained Ridge probe. It uses the observed all-layer replay contrast, which requires both conditions. Among 31 trajectories with an Opus-judged plan-alignment score below $0.5$, the first contrast crossing below $0.05$ occurs 4.45 steps before the first judged deviation on average (median 5) and occurs first in $74.2\%$ of cases. This is a retrospective diagnostic, not a deployable early-warning result.

\section{Scope Check 1: Scale and the HotpotQA Audit}
\label{sec:crossdomain}

We examine a smaller model in the same family and audit an earlier HotpotQA collection.

\textbf{Archived HotpotQA collection (excluded from validation).} The earlier runner combines plan elicitation and the first action in one assistant response, then appends that plan-bearing response to B before the second measurement. Thus B is no longer plan-omitted at the second action. Attrition also leaves 49 runs at the first action and 25 at the second. Although the stored means decrease numerically, they do not estimate a clean plan-exchange decay. We exclude this collection from the paper's validation evidence.

\textbf{Cross-scale (Llama-3.1-8B).} On 20 ALFWorld tasks, the mean contrast is $0.382$ at the first action after the plan exchange and $0.068$ at the second, a $5.6\times$ decrease. The layer profile peaks at a similar relative depth: $L_{12}$ of 32 layers ($37.5\%$) for 8B and $L_{32}$ of 80 layers ($40\%$) for 70B (Appendix~\ref{app:multimodel}).

\section{Scope Check 2: Reasoning Models}
\label{sec:reasoning}

We next test whether the Llama timing pattern extends to reasoning models. It does not do so uniformly, and standard replay pairing can obscure the difference.

\textbf{Trace contamination and control requirement.} Re-derived plan text in prior \texttt{<think>} blocks can contaminate condition B, but stripping only B also creates a distribution shift: the archived asymmetric run has pre-plan $L_{32}$ contrast $0.0199$. Five archived symmetric A--B pairs yield mean $L_{32}$ contrast $0.0309$ at the second action after the plan exchange, but the collection runner is not retained and the standard collection uses separate stochastic trajectories. We therefore report no recovery fraction and treat the symmetric value as descriptive only. Neither operator isolates plan content.

\textbf{A different temporal pattern in Qwen3.} In an exploratory native-Qwen3 dataset (15 tasks; 270 measured call pairs), the plan-exchange contrast grows to about $0.13$ by the latest measured calls rather than decreasing. This result uses a different model and longer trajectories, so it only shows that the temporal pattern is not universal; it is not a controlled model comparison.

\section{Practical Stress Test: Preserving the Plan Does Not Recover the Gap}
\label{sec:compression}

Representational change is not the same as behavioral failure. We therefore evaluate actual context compression in a separate stress test. The aggregate archive contains 150 runs per policy, described by the experiment as five runs on each of 30 ALFWorld tasks with Llama-3.1-70B. The task-level run table and launch configuration were not retained. This experiment uses a separate three-example agent prompt. The \texttt{none} condition keeps full history and serves as an uncompressed reference. The three compressed policies first retain the system prompt and four recent messages, then make policy-specific additions: \texttt{naive} adds nothing, \texttt{plan\_protected} pins the two-message plan exchange, and \texttt{probe\_gated} adds a reminder when its saved regression probe fires. Their context budgets therefore differ.

\textbf{Naive eviction has lower descriptive success; protecting the plan does not recover it.} Run-level success is $56.7\%$ without eviction and $22.0\%$ with naive eviction. Plan protection reaches $20.7\%$, and probe-gated reminders reach $24.7\%$ with an average of $6.1$ reminders per run. The local archive preserves aggregate outcomes but not the task-level run table needed for clustered inference, so we report no standard errors, confidence intervals, or $p$-values. Because naive eviction also removes observations and actions, uses a separate prompt, and changes the context budget, this result is descriptive and cannot be attributed to plan removal.

\section{Takeaways and Limitations}
\label{sec:discussion}
\label{sec:limitations}

\textbf{Main takeaway.} We show that, in the primary Llama setting, retaining the plan exchange strongly changes the first post-plan computation and has a much smaller effect one action-observation cycle later. This does not mean that the model erased the plan. It shows that the measured effect of one earlier exchange is rapidly diluted by the growing shared suffix and by any plan-derived text replayed into both histories.

\textbf{Implication for agent builders.} A fading hidden-state contrast is not permission to evict the plan. The compression stress test points in the opposite operational direction: losing history is costly, and preserving only the plan does not restore the full-history result. These findings motivate context managers that preserve or reconstruct a broader task state, including the goal, constraints, recent observations, important tool results, and unresolved subgoals. A plan reminder alone is not a complete memory policy.

\textbf{How to use replay pairing.} Replay pairing is an audit tool for asking whether retaining an exchange still changes current computation. Before using it to remove specific tokens, a study also needs content-matched controls, a design in which both conditions first see the plan and only later differ in eviction, and symmetric treatment of reasoning traces.

\textbf{Scope.} Our primary evidence covers one Llama model family on short, replay-fixed trajectories in deterministic text environments, with actions sampled at temperature $0.7$. It does not cover long-horizon browser, software-engineering, visual, or broadly stochastic environments. We rely on one LLM judge for the retrospective plan-alignment labels.

\textbf{Safety and ethics.} Context managers can silently remove plans, constraints, safety instructions, or tool schemas. Replay pairing measures representational change, not correctness or safety. It should not be used alone to decide that information is safe to drop or that an agent action is safe to execute.

\bibliographystyle{colm2026_conference}
\bibliography{references}

\appendix

\section{Implementation Details}
\label{app:probe_details}

\textbf{Models, decoding, infrastructure.} Llama-3.1-70B-Instruct \citep{llama3}, Llama-3.1-8B-Instruct, and DeepSeek-R1-Distill-Llama-70B \citep{deepseek2025r1} run with pipeline parallelism on H200 GPUs; the 70B models use eight GPUs. Qwen3-32B-thinking supports the exploratory reasoning-model analysis. The replay-pairing experiments use temperature $0.7$, top-$p$ $0.9$, and at most 512 new tokens. The retrospective lag analysis uses the same ALFWorld replay data; it does not use a trained probe. The replay runner uses the zero-shot ReAct-format system prompt in Section~\ref{sec:plan_elicit}; the separate compression harness uses three demonstrations. Hidden states are taken at the last prompt token before response generation. The 15 sampled Llama layers are $\{0,8,16,24,28,32,36,40,44,48,52,56,64,72,80\}$.

\textbf{Compute budget.} Wall-clock per experiment on $8\times$H200 (70B-class) unless noted. Paired-trajectory ALFWorld collection ($n{=}80$ tasks $\times 2$ conditions, single-pass replay with hidden-state caching) takes about 36 H200-hours. HotpotQA replay collection takes about 8 H200-hours. Reasoning-trace development runs, including the five-task symmetric rerun, take about 22 H200-hours. Qwen reasoning-model extensions take about 14 H200-hours, and the steering sweep takes about 18 H200-hours. Probe training and analysis on cached hidden states takes about 2 CPU-days. The experiments retained in this paper use about 98 H200-hours. Including preliminary and superseded runs, the full project consumed about 450--500 H200-hours on a shared internal cluster.

\textbf{Probe training.} Exploratory binary and four-class probes use logistic regression with the \texttt{lbfgs} solver, $C=1.0$, and 5-fold stratified cross-validation on normalized features. The primary continuous readout uses Ridge regression with $\alpha=1.0$ and task-grouped folds. The aggregate compression archive does not preserve the exact serialized probe bundle or firing threshold, so we make no threshold-specific claim for that policy.

\textbf{Validity controls (full numbers).} At $L_8$: real AUROC
$1.000$, shuffled $0.297$ (gap $0.703$), within-step $0.588$,
step-index $R^2$ $0.998$. At $L_{32}$: step-index $R^2$ $0.978$.
Top-10 feature dimensions at $L_8$ and $L_{32}$ are disjoint (IoU
$=0.0$).

\textbf{Graded probe per-class.} A four-way classifier at $L_{32}$ uses 5-fold cross-validation with 80 examples per class. Precision/recall/F1 are $0.988/1.000/0.994$ at probe offset $+1$, $0.876/0.888/0.882$ at $+2$, $0.728/0.738/0.733$ at $+3$, and $0.857/0.825/0.841$ at $+5$. Most errors are between adjacent offsets; one offset-$+5$ example is classified as offset $+1$.

\textbf{Strict stripping (reasoning models).} The intended symmetric operator removes all complete \texttt{<think>}$\ldots$\texttt{</think>} blocks from each prior assistant turn in both histories while retaining content after \texttt{</think>}. The five archived symmetric pairs have zero pre-plan A--B difference, but the collection runner and seed are missing. Because the standard and symmetric datasets also use separate stochastic trajectories, we report no cross-run recovery ratio and treat the symmetric result as descriptive.

\textbf{Exploratory intervention harness.} A separate re-anchoring harness uses ALFWorld's \texttt{info["won"]} field for task success. It is distinct from the compression policy and is not part of the retrospective lag result.

\textbf{Judge.} Plan-alignment scoring uses the separate judge pass with \texttt{claude-opus-4-6} at temperature 0. The retrospective lag analysis defines deviation as the first score below $0.5$ on the $[0,1]$ scale.

\textbf{Cross-model scope (negative result).} Archival hidden states from Phi-3-Medium and Qwen2.5-72B were collected in a single-run regime where step-level activations are bit-identical across re-runs of the same question. The plan-exchange contrast requires paired histories, so it is undefined on those data, and we make no claim about non-Llama architectures from existing corpora. The most informative next experiments are replay-paired collection on a mixture-of-experts model and on an RL-tuned reasoning model. Both need fresh data collection per architecture.

\subsection{Layer Profile and Task-Type Decay}
\label{app:layer}

\begin{figure}[p]
  \centering
  \begin{subfigure}[t]{0.76\linewidth}
    \includegraphics[width=\linewidth]{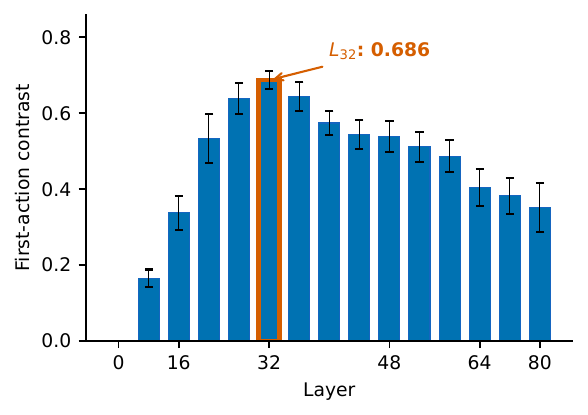}
    \caption{First-action contrast by layer.}
    \label{fig:layer_profile_app}
  \end{subfigure}
\par\medskip
  \begin{subfigure}[t]{0.90\linewidth}
    \includegraphics[width=\linewidth]{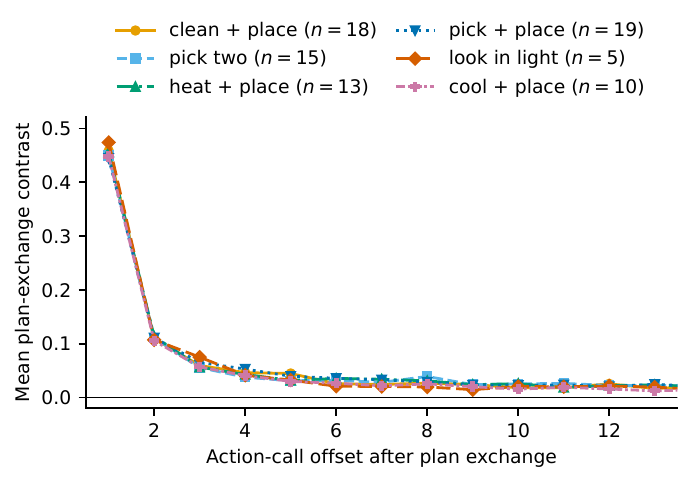}
    \caption{Mean contrast by task type and action-call offset.}
    \label{fig:task_type_app}
  \end{subfigure}
  \caption{Layer profile and task-type decay. Panel (a) reports the
    layer-specific contrast at the first action call after the plan exchange and peaks at
    $L_{32}$ ($0.686$). Panel (b) averages across all 15 sampled layers
    within each task type; all six task types show similar first-to-second-action decreases.}
\end{figure}

Figure~\ref{fig:layer_profile_app} shows the first-action layer profile across 15 sampled layers and the per-task-type decay curves. The contrast is largest at $L_{32}$ ($0.686$). All six ALFWorld task types show similar first-to-second-action decreases ($4.0$-$4.4\times$) across the plan lengths represented in this dataset.

\subsection{Probe Validity Controls and Leakage Analysis}
\label{app:leakage}

The probe dataset excludes the first action call after the plan exchange. It labels the second action call as probe offset $+1$. Binary diagnostics compare probe offset $+1$ with offsets $+3$ and later. A classifier could learn this offset rather than the contrast magnitude, so Table~\ref{tab:leakage_app} reports shuffled-label and step-index controls.

At $L_8$, the real binary AUROC is $1.000$, while the shuffled-label estimate is $0.297$.\footnote{The estimate below $0.5$ reflects variance from small-sample 5-fold cross-validation under shuffling; the relevant result is the drop from $1.000$.} A separate regressor predicts step index at $R^2=0.998$, confirming that $L_8$ carries strong step information. Step index also remains highly decodable at $L_{32}$ ($R^2=0.978$).

The primary control holds out complete tasks and compares the continuous hidden-state readout with a step-only baseline. On these held-out tasks, the probe reaches $R^2=0.847$ versus $0.575$ for the baseline, with partial $R^2=0.566$ (Section~\ref{sec:detection}).

\begin{table}[h]
\centering
\small
\begin{tabular}{lcccc}
\toprule
Layer & Real AUROC & Shuffled AUROC & $\Delta$ & Step $R^2$ \\
\midrule
$L_8$    & 1.000 & 0.297 & 0.703 & 0.998 \\
$L_{32}$ & 1.000 & n/a   & n/a   & 0.978 \\
\bottomrule
\end{tabular}
\caption{Exploratory probe-offset controls. Real: binary AUROC of offset $+1$
  vs.\ offsets $\geq+3$. Shuffled: same probe with labels permuted.
  Step $R^2$: linear regressor predicting step index from hidden
  state.}
\label{tab:leakage_app}
\end{table}

\subsection{Graded Probe Confusion Matrix}
\label{app:graded}

The four-class probe ($\{+1,+2,+3,+5\}$ probe offsets) at $L_{32}$ uses 5-fold CV on 80 examples per class. It attains accuracy $0.863$, macro $F_1=0.862$, and macro AUROC $0.976$ (Table~\ref{tab:graded_app}). Most errors are between adjacent offsets, but one offset-$+5$ example is classified as $+1$.

\begin{table}[h]
\centering
\small
\begin{tabular}{lccccc}
\toprule
True $\downarrow$ Pred $\rightarrow$ & $+1$ & $+2$ & $+3$ & $+5$ & $F_1$ \\
\midrule
$+1$ & \textbf{80} & 0  & 0  & 0  & 0.994 \\
$+2$ & 0  & \textbf{71} & 9  & 0  & 0.882 \\
$+3$ & 0  & 10 & \textbf{59} & 11 & 0.733 \\
$+5$ & 1  & 0  & 13 & \textbf{66} & 0.841 \\
\bottomrule
\end{tabular}
\caption{Four-way probe-offset classifier at $L_{32}$ (5-fold CV, 80
  examples/class). Rows are true offsets and columns are predicted offsets.}
\label{tab:graded_app}
\end{table}

\subsection{Symmetric Strict-Strip Control}
\label{app:strictstrip_table}

Across five archived R1-Distill-Llama-70B symmetric pairs, the $L_{32}$ contrasts at the second action after the plan exchange are $0.028$, $0.020$, $0.045$, $0.030$, and $0.032$ (mean $0.0309$). Their zero pre-plan difference is consistent with symmetric history construction, but the missing collection runner prevents full provenance verification. Standard and asymmetric values come from separate stochastic trajectories, so cross-run ``recovery'' percentages are not matched estimates and are omitted.

\subsection{Reasoning-Model Per-Layer Contrast}
\label{app:r1layer}

Table~\ref{tab:r1layer_app} reports the second-action contrast by layer in separate five-task R1 and Llama collections. The descriptive R1/Llama ratio ranges from $0.07$ to $0.56$, and the largest values occur at different layers ($L_{72}$ for R1 and $L_{32}$ for Llama). We do not attach inferential uncertainty to this small cross-model comparison.

\begin{table}[h]
\centering
\small
\begin{tabular}{lccc}
\toprule
Layer & R1-Distill-70B & Llama-70B & Ratio \\
\midrule
$L_{8}$  & 0.002 & 0.031 & 0.07 \\
$L_{32}$ & 0.023 & \textbf{0.184} & 0.12 \\
$L_{40}$ & 0.026 & 0.129 & 0.20 \\
$L_{48}$ & 0.030 & 0.113 & 0.27 \\
$L_{56}$ & 0.032 & 0.096 & 0.33 \\
$L_{72}$ & \textbf{0.039} & 0.070 & 0.56 \\
$L_{80}$ & 0.022 & 0.044 & 0.50 \\
\bottomrule
\end{tabular}
\caption{Descriptive per-layer plan-exchange contrast at the second action after the plan exchange in separate five-task R1-Distill-Llama and Llama collections.}
\label{tab:r1layer_app}
\end{table}

\subsection{Cross-Scale Multi-Model Panel}
\label{app:multimodel}

\begin{figure}[h]
  \centering
  \includegraphics[width=0.92\linewidth]{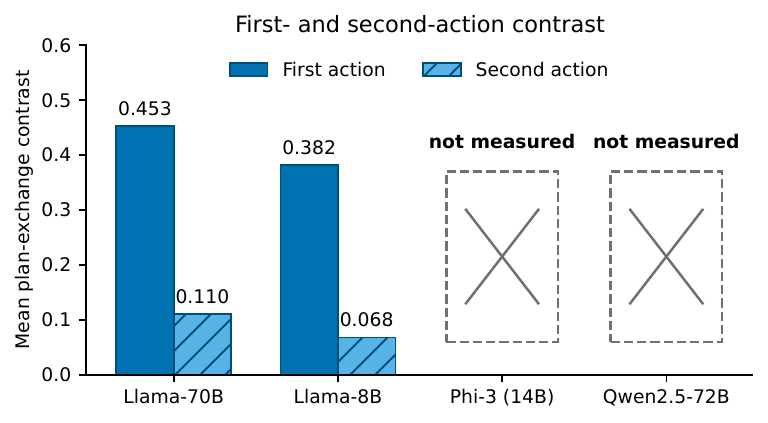}
  \caption{Cross-scale first- and second-action contrast. Llama-3.1-70B
    drops $4.1\times$ from $0.453$ to $0.110$, and Llama-3.1-8B drops
    $5.6\times$ from $0.382$ to $0.068$. The peak occurs at similar relative depth
    ($L_{12}$ of 32 layers, $37.5\%$, vs. $L_{32}$ of 80, $40\%$). Phi-3 and Qwen2.5
    are shown as not measured because their archival data are not
    replay-paired; no zero contrast is implied.}
  \label{fig:multimodel_app}
\end{figure}

Table~\ref{tab:multimodel_app} collects the cross-model decay results referenced in Section~\ref{sec:crossdomain}; Figure~\ref{fig:multimodel_app} summarizes them graphically.

\begin{table}[h]
\centering
\small
\begin{tabular}{lcccc}
\toprule
Model & First action & Second action & Decrease & Peak layer \\
\midrule
Llama-3.1-70B & 0.453 & 0.110 & $4.1\times$ & $L_{32}$ ($40\%$) \\
Llama-3.1-8B  & 0.382 & 0.068 & $5.6\times$ & $L_{12}$ ($37.5\%$) \\
Phi-3-Medium  & \multicolumn{4}{c}{\textit{N/A (not replay-paired)}} \\
Qwen2.5-72B   & \multicolumn{4}{c}{\textit{N/A (not replay-paired)}} \\
\bottomrule
\end{tabular}
\caption{Cross-scale plan-exchange contrast. Llama-3.1-70B: 80 ALFWorld tasks.
  Llama-3.1-8B: 20 tasks. Phi-3 / Qwen2.5: not replay-paired
  (archival-only data; the contrast is undefined).}
\label{tab:multimodel_app}
\end{table}

\subsection{Intervention Sweeps and Head-Level Analysis}
\label{app:intervention}

\textbf{Context-level re-anchoring.} On 30 held-out ALFWorld tasks ($\texttt{max\_steps}=200$), the control, probe-triggered, and oracle always-replan conditions all finish within one solved task of one another. This experiment shows no benefit from repeatedly re-inserting the plan text.

\textbf{Residual-stream steering ($n_{\mathrm{runs}}=5$, $n=150$ paired outcomes).} At $L_{32}$ we test $\alpha\in\{0.5,1.0,2.0\}$ on 30 held-out ALFWorld tasks with five paired runs. At $\alpha=0.5$, success changes from $54.7\%$ to $56.0\%$ ($\Delta=+1.33$pp; paired Wilcoxon $p=0.97$; 95\% bootstrap CI $[-8.0,+11.3]$). The first three runs give $+8.9$pp, while two added runs give $-10.0$pp. At $\alpha=1.0$ and $2.0$, success is $3.3$pp below baseline. We find no evidence that uniform addition along this measured direction improves task success.

\textbf{Head-level exploratory analysis.} We decompose the $L_{32}$ contrast into per-head contribution scores $c_h$. The top three heads account for $23\%$ of $\sum_h|c_h|$, and the top ten account for $55\%$; signs are mixed. A sign-aware intervention gives $+3.3$pp at $\beta=1.0$ over 90 paired outcomes and $-4.4$pp at $\beta=2.0$. Scaling to $\beta\approx5.3$ destabilizes generation and prevents normal termination. These interventions do not provide reliable evidence of a behavioral effect.

\subsection{HotpotQA Protocol Caveat}
\label{app:decay_diff}

The HotpotQA archive contains 49 replay runs at the first post-plan action but only 25 at the second. More importantly, its historical runner appends A's combined plan-and-action response to B after the first measurement, so B contains plan text before the second. Its aggregate means also average all 81 layers, whereas the ALFWorld result averages 15 sampled layers. We therefore exclude this archive from validation. A clean rerun must separate plan elicitation from action generation and retain complete histories.

\section{Context-Compression Stress Test Details}
\label{app:compression}

The aggregate report describes four context policies on 30 ALFWorld tasks (indices 50--79), with five runs per task and policy (150 runs/policy), using Llama-3.1-70B. The \texttt{none} reference keeps full history. Each compressed policy keeps the system prompt and four recent messages before adding policy-specific plan text. Success is the environment's \texttt{info["won"]} field. The local archive preserves policy-level aggregates but not the task-level run table, launch command, step cap, or exact serialized probe configuration. We therefore report descriptive pooled counts, rates, and reminder counts only.

\begin{table}[h]
\centering
\small
\begin{tabular}{lccc}
\toprule
Policy & Run success & Difference & Reminders/run \\
\midrule
\texttt{none}            & \textbf{85/150 (56.7\%)} & reference          & 0.0 \\
\texttt{naive}           & 33/150 (22.0\%)          & $-34.7$pp vs.\ none & 0.0 \\
\texttt{plan\_protected} & 31/150 (20.7\%)          & $-1.3$pp vs.\ naive & 0.0 \\
\texttt{probe\_gated}    & 37/150 (24.7\%)          & $+2.7$pp vs.\ naive & 6.1 \\
\bottomrule
\end{tabular}
\caption{Descriptive context-compression stress test (30 tasks, 5 runs/task/policy). The \texttt{none} row keeps full history. Compressed-policy context budgets differ, and task-clustered uncertainty is unavailable from the aggregate archive.}
\label{tab:compression}
\end{table}

\end{document}